\documentclass[sigconf]{acmart}
\AtBeginDocument{%
  }

\setcopyright{acmlicensed}

\copyrightyear{2025} 
\acmYear{2025} 
\setcopyright{cc}
\setcctype{by}
\acmConference[BCB '25]{Proceedings of the 16th ACM International Conference
on Bioinformatics, Computational Biology, and Health Informatics}{October
11--15, 2025}{Philadelphia, PA, USA}
\acmBooktitle{Proceedings of the 16th ACM International Conference on
Bioinformatics, Computational Biology, and Health Informatics (BCB '25), October
11--15, 2025, Philadelphia, PA, USA}
\acmDOI{10.1145/3765612.3767193}
\acmISBN{979-8-4007-2200-4/2025/10}

\usepackage{graphicx}
\usepackage{hyperref}
\usepackage{url}
\usepackage{wrapfig}
\usepackage{amsmath}
\usepackage{subfig}
\usepackage{subcaption}
\usepackage{booktabs}
\usepackage{tabularx}
\usepackage{array}
\usepackage{multirow}
\usepackage{enumitem}

\begin{document}

\title{Retrieval-Reasoning Large Language Model-based Synthetic Clinical Trial Generation}

\author{Zerui Xu}
\affiliation{%
  \institution{University of Chicago}
  \city{Chicago}
  \country{USA}}
\email{xuzr3x@uchicago.edu}

\author{Fang Wu}
\affiliation{%
  \institution{Stanford University}
  \city{Stanford}
  \country{USA}}
\email{fangwu97@stanford.edu}

\author{Yingzhou Lu}
\affiliation{%
  \institution{Stanford University}
  \city{Stanford}
  \country{USA}}
\email{lyz66@stanford.edu}

\author{Yuanyuan Zhang}
\affiliation{%
  \institution{Purdue University}
  \city{West Lafayette}
  \country{USA}}
\email{zhang038@purdue.edu}

\author{Yue Zhao}
\affiliation{%
  \institution{University of Southern California}
  \city{Los Angeles}
  \country{USA}}
\email{yzhao010@usc.edu}

\renewcommand{\shortauthors}{Xu et al.}

\begin{abstract}
Machine learning (ML) holds great promise for clinical applications but is often hindered by limited access to high-quality data due to privacy concerns, high costs, and long timelines associated with clinical trials. While large language models (LLMs) have demonstrated strong performance in general-purpose generation tasks, their application to synthesizing realistic clinical trials remains underexplored. In this work, we propose a novel \textit{Retrieval-Reasoning} framework that leverages few-shot prompting with LLMs to generate synthetic clinical trial reports annotated with binary success/failure outcomes. Our approach integrates a retrieval module to ground the generation on relevant trial data and a reasoning module to ensure domain-consistent justifications. Experiments conducted on real clinical trials from the ClinicalTrials.gov database demonstrate that the generated synthetic trials effectively augment real datasets. Fine-tuning a BioBERT classifier on synthetic data, real data, or their combination shows that hybrid fine-tuning leads to improved performance on clinical trial outcome prediction tasks. Our results suggest that LLM-based synthetic data can serve as a powerful tool for privacy-preserving data augmentation in clinical research. The code is available at~\url{https://github.com/XuZR3x/Retrieval_Reasoning_Clinical_Trial_Generation}.
\end{abstract}



\maketitle

\section{Introduction}
One promising application of Large Language Models (LLMs) is the generation of synthetic data. This capability is especially important in domains where real data are scarce, sensitive, or expensive to acquire—such as law, finance, and notably healthcare~\cite{xu2020generating}. In medical contexts, the privacy of patient records and strict regulations around data sharing present significant obstacles to building large, labeled datasets. These challenges are particularly acute for clinical trials, which are essential for validating new treatments yet are constrained by limited availability, high costs, and long durations~\cite{bhatt2021clinical, chen2021ethical}. The scarcity of accessible clinical trial data hampers the development of machine learning (ML) models for tasks such as trial outcome prediction, patient stratification, or eligibility screening~\cite{rajkomar2019machine}.

Synthetic clinical trial data generated by LLMs offers a compelling alternative to address this challenge. By simulating realistic clinical trial narratives, researchers can create artificial datasets that preserve the statistical and structural properties of real trials while avoiding privacy concerns~\cite{jordon2020synthetic, tucker2020generating}. Such synthetic datasets can support diverse use cases: from benchmarking algorithms and training predictive models to expanding domain generalization by injecting variation and novel configurations not observed in limited real-world samples.

\begin{figure*}[t]
  \centering
  \includegraphics[width=\linewidth]{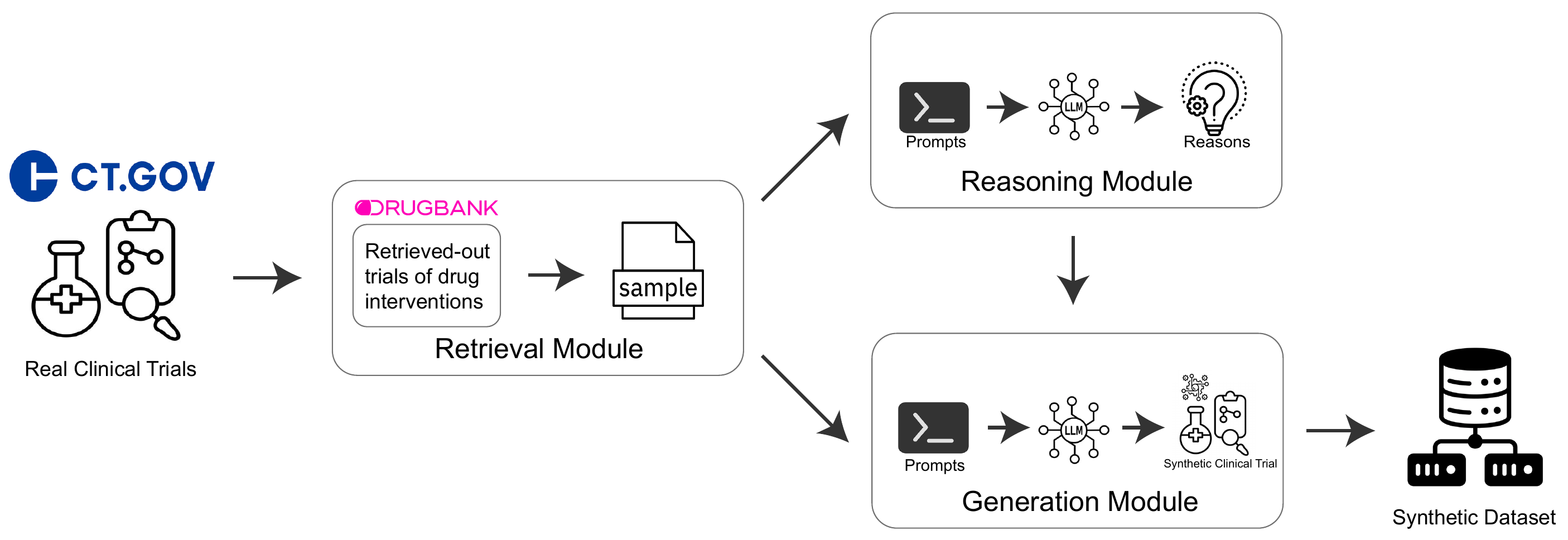}
  \caption{The overall pipeline of retrieval–reasoning clinical trial generation.}
  \Description{A three-stage pipeline diagram: (1) Retrieval: a query encoder takes a clinical question and retrieves top-k relevant trial documents. (2) Reasoning: a module extracts eligibility criteria and synthesizes components like objectives and methods. (3) Generation: a schema-driven generator produces a finalized clinical trial protocol with title, eligibility, interventions, and outcomes.}
  \label{Main-fig}
\end{figure*}

However, generating high-fidelity clinical trial data is far from trivial. Real trial documents encode complex interactions among multiple variables (interventions, populations, outcomes, etc.), and an LLM must not only replicate the linguistic style and document structure of trials but also reflect clinically meaningful correlations that drive trial outcomes~\cite{chen2021synthetic}. Without careful guidance, LLMs risk producing plausible-sounding but medically inaccurate or logically inconsistent trial reports, limiting their utility. A further challenge is the distributional gap between synthetic and real data~\cite{xu2020generating}. While synthetic data can be scalable and diverse, it may lack the subtle patterns and complex dependencies of authentic clinical records. This discrepancy can degrade model performance when deployed on real-world tasks. In critical applications like clinical trial outcome prediction, where both false positives and false negatives carry serious consequences~\cite{chen2021ethical, rajkomar2019machine}, it is essential that synthetic data align well with real-world distributions to support reliable clinical inference.

To overcome these challenges, we propose a novel \textit{Retrieval-Reasoning} generation framework that leverages few-shot prompting with LLMs to synthesize realistic clinical trial reports annotated with binary success/failure outcomes. Our approach introduces a retrieval module that grounds generation in real trial data for known drug interventions, and a reasoning module that composes interpretable rationales explaining the outcome label. These modules work in concert to constrain the LLM and improve factual consistency in the generated trials. Central to our framework is a hybrid fine-tuning strategy that combines synthetic trials with real data to train a BioBERT classifier~\citep{lee2020biobert}. This hybrid approach bridges the domain gap by leveraging the volume, diversity, and controllability of synthetic data while grounding the model in authentic clinical patterns from real trials~\cite{jordon2020synthetic}. In doing so, we harness the complementary strengths of synthetic and real data for robust clinical trial modeling under data-scarce conditions.

\begin{figure*}[t]
  \centering
  \includegraphics[width=0.8\textwidth]{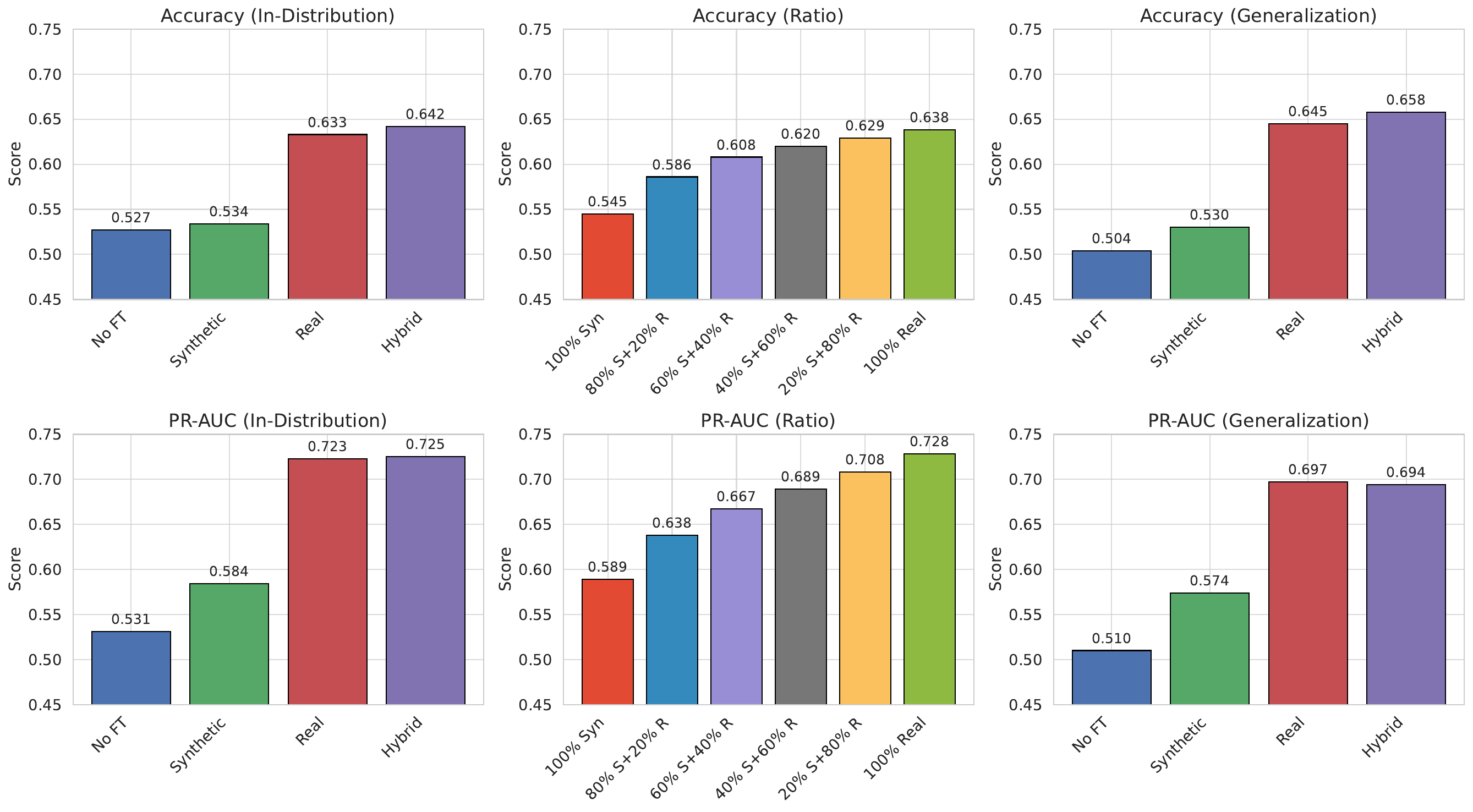}
  \caption{BioBERT performance across fine-tuning strategies. 
    \textbf{Top row:} Accuracy under in-distribution, ratio, and generalization tests. 
    \textbf{Bottom row:} PR-AUC under the same settings. 
    Each bar is colored consistently with its corresponding strategy on the x-axis.
  }
  \label{fig:performance}
\end{figure*}

Our main contributions are three-fold:
\begin{itemize}[leftmargin=*]
    \item We develop a retrieval–reasoning pipeline (see Fig.~\ref{Main-fig}) to generate diverse, high-fidelity synthetic clinical trial reports with explicit outcome labels, providing an effective data augmentation strategy for clinical ML tasks.
    \item We demonstrate that hybrid fine-tuning on both real and synthetic trials significantly improves classification performance for trial outcome prediction, especially in data-scarce settings (see Fig.~\ref{fig:performance}).
    \item Through t-SNE visualization and cosine similarity analysis, we show that our synthetic data enriches the feature space, broadening coverage and improving model robustness by complementing real trial data.
\end{itemize}
Collectively, our work highlights the promise of LLM-based synthetic trial generation as a tool for scalable, privacy-preserving clinical research.

\section{Related Work}

\paragraph{Synthetic Data Generation}
AI-driven synthetic data generation provides a way to overcome data scarcity and privacy constraints by creating artificial yet realistic datasets. In healthcare, patient information is commonly stored in electronic health records (EHRs)~\citep{kruse2017security,wen2022disentangled,fu2019pearl}, which have been widely leveraged for medical research~\citep{goncalves2020generation,du2023abds}. MedGAN~\citep{choi2017generating}, combining autoencoders and generative adversarial networks, generates high-dimensional discrete patient records and has shown strong performance across distributional statistics, downstream classification tasks~\citep{lu2018multi}, and expert evaluations~\citep{bhanot2021problem,fu2024ddn3,zhang2021ddn2,chen2021data,choi2017generating}. Synthetic data also mitigates regulatory barriers to data sharing across organizations~\citep{eigenschink2021deep,wu2022cosbin}, enabling broader collaboration and reducing biases in downstream studies~\citep{tucker2020generating,wang2022eeg,wu2022cosbin}.

\paragraph{AI for Clinical Trials}
Deep learning has been applied to a variety of clinical trial tasks, including outcome and approval prediction~\citep{fu2022hint,fu2023automated,lu2024uncertainty,yue2024ct,tang2025medagentsbench}, duration estimation~\citep{yue2024trialdura}, enrollment success prediction~\citep{yue2024trialenroll}, patient dropout modeling~\citep{chen2024trialbench}, and digital twin simulation~\citep{wang2024twin}. Clinical trials remain costly~\citep{martin2017much} and time-consuming~\citep{peto1978clinical,ledford20114}, with failures often caused by drug ineffectiveness, safety issues, or poorly designed eligibility criteria~\citep{friedman2015fundamentals}. These challenges motivate outcome prediction approaches to reduce unsuccessful trials and allocate resources more efficiently~\citep{lu2024uncertainty,lu2022cot}.

\section{Method}

\subsection{Preliminaries and Background}  
\paragraph{Data Source.} \url{ClinicalTrials.gov} is a comprehensive public database maintained by the U.S. National Library of Medicine (NLM) that provides detailed information about clinical trials worldwide. Each entry typically contains the study’s purpose, design, eligibility criteria, locations, and outcomes. We use the entire set of trials from \url{ClinicalTrials.gov} as our real dataset. These trial reports (originally in XML format) are converted into text strings $S_i$, and the complete collection is $\mathbb{S}_{\text{total}} = \{S_i\}_{i=1}^{N_{\text{total}}}$, where $N_{\text{total}} = 494,290$. Within this collection, a subset of trials has been annotated with binary outcomes by a team at IQVIA, with $y_j \in \{0,1\}$ indicating failure or success. Let $\mathbb{S}_{\text{labeled}} \subset \mathbb{S}_{\text{total}}$ denote the set of labeled trials, of size $N_{\text{labeled}} = 26,768$. We use the labeled pairs $\{(S_j, y_j)\}_{j=1}^{N_{\text{labeled}}}$ for model training and evaluation.

\paragraph{Few-Shot In-Context Generation with LLMs.} We leverage an LLM $\mathcal{M}$ (ChatGPT-4o-mini) to perform few-shot in-context learning. In this paradigm, the model is provided with a prompt $\mathcal{P}$ containing a few example input-output pairs and then asked to generate an output for a new input without updating its parameters. Formally, given $K$ examples $\{(I_i, O_i)\}_{i=1}^K$ and a new input $I_{K+1}$, we construct a prompt 
$\mathcal{P} =$ \texttt{Concat}($\text{format}(I_1,O_1), \ldots, \text{format}(I_K,O_K), 
\\
\text{format}(I_{K+1})$), where each $\text{format}(I_i,O_i)$ represents how the example is presented. The model then generates an output $O_{K+1}$ by modeling the conditional probability $P_{\theta}(O_{K+1} \mid \mathcal{P}) = \prod_{t=1}^{T} P_{\theta}(o_t \mid \mathcal{P}, o_{<t})$, where $O_{K+1}=(o_1,\dots,o_T)$ is the token sequence of the output. In essence, at each step the LLM chooses the next token $o_t$ based on the prompt and all previously generated tokens $o_{<t}$. This allows the model to learn the task demonstrated by the examples (via in-context meta-learning) and produce an output $O_{K+1}$ that aligns with the style and structure of those examples.

\paragraph{Fine-Tuning for Outcome Prediction.} We fine-tune a pretrained BioBERT model to classify clinical trial outcomes. Given a clinical trial report represented as a feature vector $\mathbf{x}_i$, the model learns a function $f_\theta(\mathbf{x}_i)$ that outputs a predicted label $\hat{y}_i \in \{0,1\}$. We train $f_\theta$ on a labeled dataset $\{(\mathbf{x}_i, y_i)\}_{i=1}^N$ by minimizing the binary cross-entropy loss. During inference, the model produces a probability $\hat{p} = P_\theta(y=1 \mid \mathbf{x}_{\text{new}})$ for a new trial $\mathbf{x}_{\text{new}}$. We classify the trial as a success ($\hat{y}=1$) if $\hat{p} \ge 0.5$ and failure ($\hat{y}=0$) otherwise. We evaluate performance using metrics such as accuracy, precision, recall, ROC-AUC, and PR-AUC.

\subsection{Retrieval-Reasoning Few-Shot Generation}

\subsubsection{Overview}
Our generation framework (Fig.~\ref{Main-fig}) consists of three modules: \textit{retrieval}, \textit{reasoning}, and \textit{generation}. We modify the standard few-shot prompting approach to better control the output. Instead of prompting the LLM with an input to predict an output, we provide the desired outcome label and task the LLM with generating a clinical trial report that would produce that outcome. Specifically, we select $K=3$ example trials $\{(S_i, y)\}_{i=1}^3$ that share the same intervention (drug) and the same outcome label $y$. We then construct a single composite prompt for $\mathcal{M}$ that includes: (i) a context setting (instructing the model to act as a medical expert), (ii) a list of $K$ retrieved example trials $(S_i, y)$ as in-context examples, (iii) a set of reasons $\mathcal{R}$ explaining why those trials had outcome $y$ (generated by the reasoning module described below), (iv) a formatting constraint (to ensure the output follows the structure of a trial report), and (v) an instruction to generate a new trial report $S_{\text{new}}$ with the specified outcome $y$, along with a final diversity prompt encouraging uniqueness. By grounding the prompt in real examples and explicit rationale, the LLM is guided to produce a plausible trial report consistent with the given outcome.

\subsubsection{Retrieval Module} 
The retrieval module filters the pool of real trials to find candidates involving well-known drug interventions, which helps ground the generation in a realistic medical context. In particular, we identify drug names from the DrugBank database (\url{https://www.drugbank.ca/}) and retrieve clinical trials that involve those drugs. We further require that the chosen intervention have at least three successful trials and three failed trials in the labeled dataset. This ensures we can sample three example trials for the prompt that share the same intervention and outcome $y$ (either all successes or all failures). Using widely recognized drug interventions makes it more likely that the LLM will generate trials about realistic treatments rather than obscure or implausible ones. Once an intervention and outcome $y$ are fixed, we sample three corresponding trials $\{(S_i, y)\}_{i=1}^3$ to use as few-shot examples in the subsequent modules.

\subsubsection{Reasoning Module} 
Given an intervention and outcome $y$, the goal of the reasoning module is to generate a set of plausible reasons explaining why trials with this intervention resulted in success or failure. We prompt the LLM to produce five distinct reasons. The prompt for this module includes a brief context (positioning the LLM as a medical expert analyzing trials of the intervention), the three example trials $(S_i, y)$ retrieved earlier (with the outcome label explicitly noted), a directive that the output should be exactly five enumerated points, and an instruction to "Write 5 reasons why the trials of [intervention] would have [succeeded/failed]." We also add a final prompt asking for more diverse content to encourage variety in the reasons. The LLM then outputs five concise reasons (numbered 1 through 5) that could explain the given outcome. This reasoning step provides interpretable factors (e.g., efficacy of the drug, trial design issues, patient population differences) that can guide the narrative of the generated trial. See Supplementary Materials for the complete reasoning prompt template.

\subsubsection{Generation Module}
In the generation module, we combine all the pieces to produce a new synthetic trial report. The prompt to the LLM consists of: the context instruction (medical expert persona), the list of five reasons from the reasoning module (highlighting key factors for success or failure), the three example trials $(S_i, y)$ as templates and references, a constraint that the output should follow the format and style of a ClinicalTrials.gov report (an XML-like structured format as seen in the examples), and finally an explicit instruction to "Write a report of a [successful/failed] clinical trial of [intervention]" along with a diversity encouragement. This structured prompt ensures that the generated report is both format-compliant and factually supported by the reasons and examples provided. Using this pipeline with ChatGPT-4o-mini (temperature set to 1), we generated a total of 3,358 synthetic clinical trial reports, each annotated with a binary outcome. We recorded the intervention name and outcome for each generated trial. The exact prompt structure for the generation module is also provided in Supplementary Materials.

\section{Experiments}
We conducted a series of experiments to evaluate the effectiveness of our synthetic clinical trials for improving outcome prediction models. We fine-tuned a pretrained BioBERT model as a classifier under various training data settings and assessed its performance on both in-distribution and out-of-distribution scenarios. In addition, we visualized the representation spaces of real and synthetic trials using t-SNE, and measured the diversity within and between these sets using cosine similarity. These analyses illustrate how well the synthetic data aligns with real data and how it broadens the training distribution.

\begin{figure*}[t]
  \centering
  \includegraphics[width=0.48\linewidth]{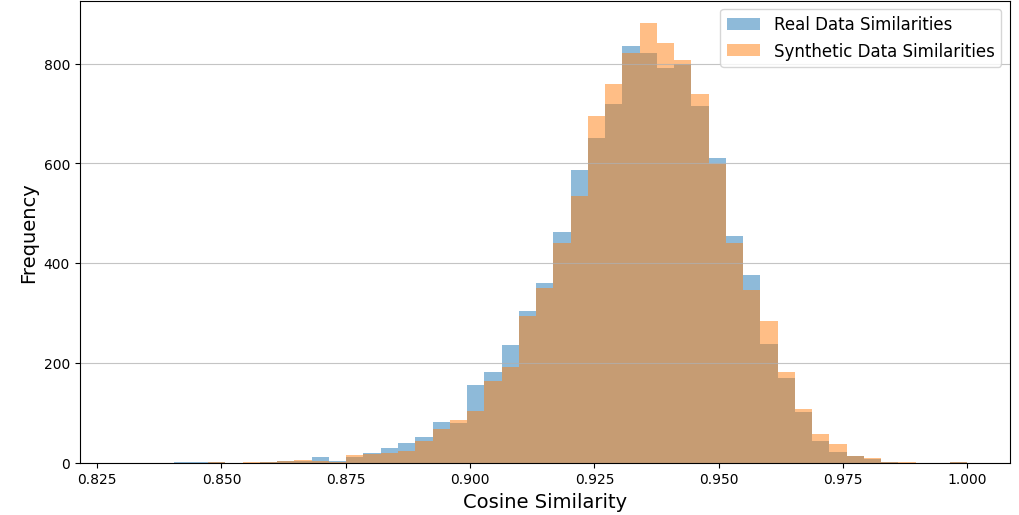}\hfill
  \includegraphics[width=0.48\linewidth]{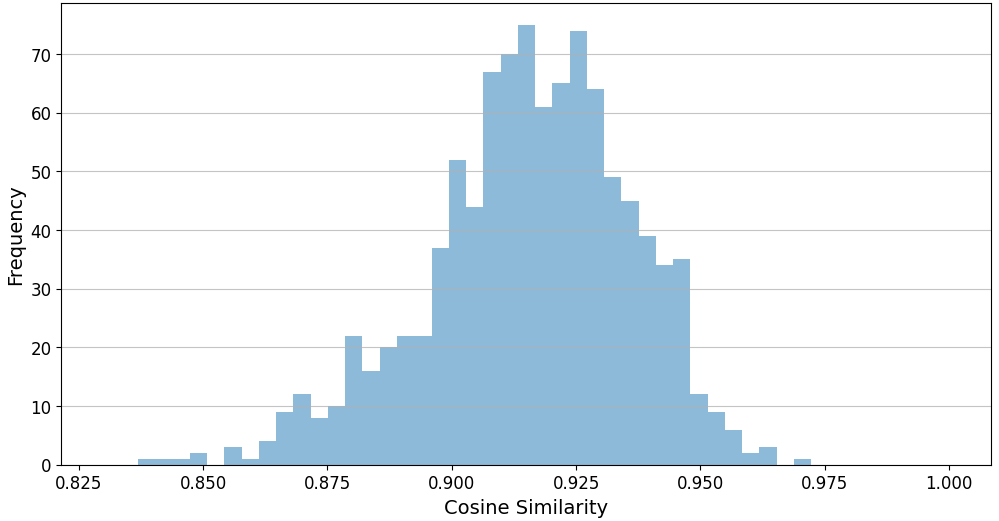}
  \caption{Cosine similarity distribution of trial embeddings. \textbf{Left:} Similarity among random pairs within the real dataset and within the synthetic dataset. \textbf{Right:} Similarity among random real-synthetic trial pairs. Synthetic trials exhibit high internal consistency comparable to real trials, while the cross-distribution similarities are more varied, indicating that some synthetic trials closely resemble real ones, whereas others are more novel.}
  \label{fig:cossim}
\end{figure*}

\begin{figure}[t]
  \centering
  \includegraphics[width=\columnwidth]{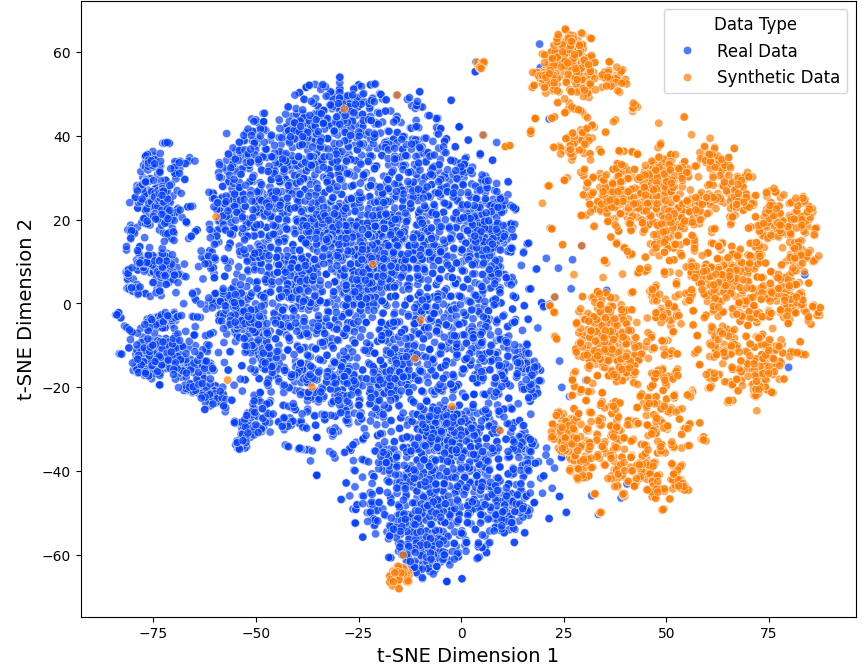}
  \caption{t-SNE visualization of real vs. synthetic trial embeddings. Blue points are real trials and orange points are synthetic. The plot shows that synthetic trials overlap with real trials in many areas of the representation space, while also covering additional areas (note the spread of orange points), thereby enriching the overall data distribution.}
  \label{fig:t-SNE}
\end{figure}

\subsection{Fine-Tuning BioBERT for Outcome Classification}
We assess the contribution of synthetic data by fine-tuning BioBERT under three settings: using only synthetic trials, using only real trials, and using a hybrid mix of both. We first partition the real labeled dataset based on whether the trial’s intervention appears in our synthetic set. Let $\mathbb{A}$ be the set of labeled real trials whose intervention is included among the generated synthetic trials, and $\mathbb{B}$ be the complementary set of trials with other interventions. (We found $|\mathbb{A}|=6,056$ and $|\mathbb{B}|=20,712$.) 

For the \textbf{in-distribution} evaluation, we trained and tested on trials involving interventions seen in the synthetic data. We split $\mathbb{A}$ into training (60\%), validation (20\%), and test (20\%) sets. This yielded 3,633 real training trials, and 1,211/1,212 trials for validation/test respectively. We then created three training sets: (1) \textit{Synthetic-Only}, consisting of 3,358 synthetic trials (the full set of generated data); (2) \textit{Real-Only}, consisting of the 3,633 real trials from $\mathbb{A}$; and (3) \textit{Hybrid}, combining all synthetic and real trials (3,358 + 3,633 = 6,991 total). 

We also conducted a \textbf{ratio experiment} to examine performance as the proportion of real vs. synthetic data varies. In this setting, we fixed the training set size at 3,358 samples and created six mixes: 100\% synthetic, 80\% synthetic + 20\% real, 60\% synthetic + 40\% real, 40\% synthetic + 60\% real, 20\% synthetic + 80\% real, and 100\% real. The real portions were drawn from $\mathbb{A}$. The validation and test sets (1,349 trials each) were also drawn from $\mathbb{A}$.

For the \textbf{out-of-distribution generalization} test, we trained on $\mathbb{A}$ but evaluated on $\mathbb{B}$ (trials with interventions that were never seen in the synthetic data). We balanced $\mathbb{B}$ by downsampling the majority class, yielding 7,546 trials for validation and 7,546 for test. The training sets in this case were: Synthetic-Only (3,358 synthetic trials), Real-Only (all 6,056 trials in $\mathbb{A}$), and Hybrid (combined 9,414 trials). 

We fine-tuned the BioBERT base model (110M parameters) for 7 epochs in each setting (learning rate $1\times10^{-5}$, batch size 8). To prevent label leakage, we removed any explicit outcome indicators from the text of the trials (e.g., phrases in the trial description like "Overall Status: Completed" or "Why stopped" reasons for terminated trials, as well as any occurrence of the words “successful” or “failed” in synthetic trials). Each experiment was repeated three times with random seeds (40, 41, 42) for robustness. We report the mean and standard deviation of accuracy, precision, recall, ROC-AUC, and PR-AUC on the test sets. Full numeric results for all evaluation metrics are available in Supplementary Materials.

\subsection{Ablation Studies}
To assess the contribution of each module in our \textit{Retrieval-Reasoning} pipeline, 
we performed ablation experiments by removing either the retrieval module or the reasoning module 
during synthetic trial generation. We then fine-tuned BioBERT on these ablated datasets.

Table~\ref{tab:ablation} shows that removing the retrieval module causes a clear performance drop 
(accuracy from 65.73\% to 63.20\%), highlighting the importance of grounding in real trial data. 
In contrast, removing the reasoning module results in only a minor change, 
suggesting that explicit rationales aid interpretability more than quantitative gains.

We also examined the robustness of our method to prompt variations (see Supplementary Materials). The results show minimal impact on performance when altering outcome label wording or example order, indicating the framework’s prompts are not overly sensitive to these choices.

\begin{table}[t]
\centering
\caption{Ablation of pipeline modules. Downstream classifier performance with synthetic data generated under different settings.}
\label{tab:ablation}
\begin{tabular}{lcc}
\toprule
\textbf{Training Data Setting} & \textbf{Accuracy (\%)} & \textbf{PR-AUC} \\
\midrule
Full pipeline (Retrieval+Reasoning) & \textbf{65.73} & \textbf{0.7800} \\
-- w/o Retrieval module             & 63.20          & 0.7747 \\
-- w/o Reasoning module             & 65.33          & 0.7795 \\
\bottomrule
\end{tabular}
\end{table}

\subsection{Representation Analysis}
To assess how synthetic data complements real data, we analyze BioBERT embeddings. We compute cosine-similarity distributions for 10{,}000 random pairs within each set (real–real, synthetic–synthetic) and across sets (real–synthetic). As shown in Figure~\ref{fig:cossim}, intra-set curves concentrate at higher similarity, indicating internal coherence. The real–synthetic curve is broader and lower on average, implying a mix of close overlaps and genuinely novel combinations absent from the real corpus. This overlap-plus-novelty pattern is desirable for augmentation, preserving clinical signals while adding diversity that can improve robustness.

We also project embeddings with t-SNE (Figure~\ref{fig:t-SNE}). Real and synthetic trials form two \emph{distinct} clusters with only minimal overlap, indicating a distributional gap between generated and real data. The synthetic cluster shows a slightly wider spread, consistent with expanded coverage and additional modes introduced by generation. Together with the cosine-similarity results, this suggests that synthetic trials \emph{complement rather than replicate} real data: they broaden coverage while real data anchor the decision boundary—explaining the generalization gains of hybrid training (Figure~\ref{fig:performance}).

\subsection{Qualitative Audit of Synthetic Trials}
We performed a manual qualitative audit using a simple checklist covering objectives, endpoints, adverse events, rationale, and overall structure. Generated trials were generally found to be plausible and well-structured, mirroring common patterns in real-world clinical studies. Full exemplars and the complete checklist are provided in the Supplementary Materials.

\section{Conclusion} 
We introduced a framework that leverages retrieval-augmented, reasoning-guided LLM generation to produce realistic synthetic clinical trial reports labeled with outcomes. By addressing data scarcity and privacy constraints, these synthetic trials enable model training and evaluation without relying solely on real patient data. Our experiments demonstrated that augmenting real data with LLM-generated trials can improve a model's performance in predicting trial outcomes, especially in low-data settings. The synthetic data also accelerates experimentation and can facilitate collaborative research by sidestepping privacy barriers. Overall, our results highlight the promise of LLM-based synthetic data generation in healthcare, particularly for enhancing machine learning applications when real-world data are limited or sensitive. In future work, we plan to explore integrating additional modalities (e.g., structured data or imaging) and extending our approach to more complex trial scenarios and endpoints.

\section*{Limitations}
The quality of our synthetic trials is inherently tied to the LLM used; any biases or inaccuracies in the LLM’s training data may be reflected in the generated text. While our synthetic data provides useful augmentation to real trials, the t-SNE analysis reveals a distributional gap between the two, indicating that synthetic trials are not perfect replicas of real ones. The performance differences observed in the ratio experiments further highlight the need to narrow this gap and improve the fidelity of generated trials. Additionally, our study focuses only on drug intervention trials, which represent a subset of all clinical trials. Other types of interventions (e.g., surgical or behavioral studies) were not explored and may require different generation strategies. For a detailed error analysis of the generated trial reports – including common rationale themes and minor artifacts/hallucinations observed – please refer to Supplementary Materials.

\section*{Ethical Considerations}
\paragraph{Potential Risks.} Using LLMs to generate clinical trial data carries potential risks. Biases present in the LLM’s training data could lead to synthetic trials that misrepresent certain patient populations or treatment outcomes. Furthermore, over-reliance on synthetic data might reduce the robustness of models if important real-world complexities are not captured by the generated trials. It is crucial to continually validate model performance on genuine clinical data and to use synthetic data as a supplement rather than a replacement for real-world evidence.

\bibliographystyle{ACM-Reference-Format}
\bibliography{custom}

\appendix
\renewcommand{\appendixname}{Supplementary Material}

\section{Supplementary Materials}
\label{sec:additional_results}

\subsection{Prompt Design and Templates}

\subsubsection{Overview}
We designed a suite of prompts to guide ChatGPT-4o-mini in producing clinically plausible and well-structured trial reports. The design followed five principles: (i) providing a clear clinical context, (ii) including real trial exemplars, (iii) enforcing constraints on format and structure, (iv) encouraging diversity in outputs, and (v) linking reasoning directly to trial outcomes. Together, these components ensured outputs aligned with clinical scenarios and supported consistent generation at scale.

\subsubsection{Context Prompt}
Establishing a clear background framework is crucial for guiding the model to produce specialized and relevant responses. Positioning the model as a medical expert directs its analytical focus toward the complexities of the medical field. This ensures that the content aligns with clinical scenarios and reflects the technical precision typical of medical language. Providing specific examples and clearly defining the task further enhance the model's ability to generate accurate and contextually appropriate responses.

\subsubsection{Example Prompt}
Incorporating real clinical trial reports as few-shot examples is key to improving the model’s reasoning abilities. For each generation, the examples \( \{(S_i, y)\}_{i=1}^3 \) with a fixed intervention name and outcome label are randomly drawn and integrated into the prompt as strings. Given that ChatGPT-4o mini has a maximum input capacity of 128K tokens, a check condition is applied to ensure that the total token count of the selected samples stays within this limit. The label $y$ is explicitly stated each time before the associated content to remind the model of the outcome.

\subsubsection{Constraint Prompt}
Imposing structural guidelines enhances the clarity and consistency of the model's output. A predefined format ensures organized and interpretable responses, facilitating comparative analysis and uniformity across outputs. This structure reduces irrelevant content and strengthens the coherence of the generated explanations, aligning them with the desired analytical framework. As a result, the model produces logically sound and contextually relevant responses within the clinical domain.

\subsubsection{Generation Prompt}
Specifying a fixed number of reasons improves the precision and comprehensiveness of the model’s output. By setting this limit, the system ensures responses are neither too brief nor overly detailed, balancing the analysis effectively. Encouraging originality in the generated reasons broadens the range of factors considered, enriching the analysis and reducing redundancy. This approach allows the model to capture the multifaceted nature of clinical trials, offering deeper insights for evaluation and decision-making.

\subsubsection{Diversity Prompt}
Promoting diverse reasoning is crucial to avoid repetitive explanations and ensure a thorough exploration of factors influencing trial outcomes. Encouraging variability in the model’s responses broadens the scope of analysis, identifying a wider range of influences and perspectives. This diversity-oriented approach enhances the quality and depth of the generated reasons, contributing to a more comprehensive and nuanced understanding of clinical interventions and their outcomes.

\subsubsection{Reasoning Prompt}
The Reasoning Prompt ensures the model's output is precisely aligned with the clinical scenario by specifying the intervention and outcome $y$. This prompt directs the model's reasoning toward relevant factors. Incorporating the five previously generated reasons provides a consistent foundation for further explanation, reinforcing logical continuity and coherence.

\subsubsection{Exact Prompt Templates}
Tables~\ref{reasoning_prompts} and~\ref{generation_prompts} provide the full prompt templates used for reasoning and generation modules. These prompts were combined with retrieval-based trial selection to generate a total of 3,358 synthetic clinical trial reports using ChatGPT-4o-mini (temperature=1). Intervention names were recorded alongside outputs for later analysis.

\begin{table}[t]
\centering
\caption{Prompts used for reasoning module.}
\label{reasoning_prompts}
\small
\begin{tabularx}{\linewidth}{p{1.5cm} X} \toprule
\textbf{Category} & \textbf{Prompt} \\ \midrule
Context & You are now a medical expert in the clinical area. You are given information about a medical intervention, and three clinical trial reports of it, either all successful or all failed. You are asked to analyze this input and write reasons resulting in the trials' success/failure. Your writing style must be consistent within the clinical study. You must ensure that your language is precise, technical, and reasonable. \\ \midrule
Example & (Successful/Failed) clinical trial example: .... \\ \midrule
Constraint & Your output should strictly follow the following format: 1. (...) 2. (...) 3. (...) 4. (...) 5. (...), with (...) being the reasons you write. \\ \midrule
Generation & Write 5 reasons leading (intervention name) to (succeed/fail) in these trials. Be creative and write unique reasons. \\ \midrule
Diversity & Can you provide something more diverse compared to the previously generated reasons? \\
\bottomrule
\end{tabularx}
\end{table}

\begin{table}[t]
\centering
\caption{Prompts used for generation module.}
\label{generation_prompts}
\small
\begin{tabularx}{\linewidth}{p{1.5cm} X} \toprule
\textbf{Category} & \textbf{Prompt} \\ \midrule
Context & You are now a medical expert in the clinical area. You are asked to write a report of a successful or failed clinical trial. Your writing style must be consistent within the clinical study. Ensure your language is precise, technical, and formal. \\ \midrule
Reasoning & Here are five reasons that could lead to the (success/failure) of clinical trials of (intervention name): ... \\ \midrule
Example & (Successful/Failed) clinical trial example: .... \\ \midrule
Constraint & Your output style should strictly follow the XML-like format of the provided examples. You cannot simply modify or rewrite them. The intervention name must be (intervention name), and you must refer to these reasons when writing clinical trials. \\ \midrule
Generation & Write a report of a (successful/failed) clinical trial of (intervention name). Ensure your language is precise, technical, and formal. Be creative and write unique reports. \\ \midrule
Diversity & Can you provide something more diverse compared to the previously generated reports? \\
\bottomrule
\end{tabularx}
\end{table}

\subsection{Prompt Sensitivity}
Prompting choices can influence LLM behavior. We therefore changed label
phrasing and example ordering to test robustness. We used the following
setting: \emph{2000} training samples (1600 real + 400 synthetic), validation/test
of 250/250 real samples (drawn from the generalized pool without intervention
restriction). We fine-tuned BioBERT and report 3-run mean $\pm$ SD.

\begin{table}[t]
\centering
\caption{Prompt-format ablations (3-run mean $\pm$ SD) in the rebuttal setting. 
Default uses ``success/failure'' labels with grouped examples.}
\label{tab:prompt_ablation}
\renewcommand{\arraystretch}{1.15}
\small
\begin{tabularx}{\linewidth}{l *{3}{>{\centering\arraybackslash}X}}
\toprule
\textbf{Prompt Variant} & \textbf{Accuracy} & \textbf{PR-AUC} & \textbf{Recall} \\
\midrule
Default (success/failure, grouped) & \textbf{0.6573 $\pm$ 0.018} & \textbf{0.7800 $\pm$ 0.007} & 0.7303 $\pm$ 0.027 \\
Positive/Negative, grouped        & 0.6400 $\pm$ 0.006          & 0.7690 $\pm$ 0.027          & 0.6962 $\pm$ 0.022 \\
Mixed outcomes, label at start    & 0.6427 $\pm$ 0.019          & 0.7624 $\pm$ 0.023          & \textbf{0.7825 $\pm$ 0.034} \\
Mixed outcomes, label at end      & 0.6373 $\pm$ 0.025          & 0.7714 $\pm$ 0.010          & 0.6888 $\pm$ 0.046 \\
Grouped, examples shuffled        & 0.6520 $\pm$ 0.015          & 0.7660 $\pm$ 0.010          & 0.7608 $\pm$ 0.058 \\
\bottomrule
\end{tabularx}
\end{table}

\paragraph{Robustness.}
All variants fall within a narrow PR-AUC band. 
\\($\approx$0.762–0.780).
The default prompt is best on PR-AUC and accuracy, but no setting
collapses, indicating limited sensitivity to phrasing or ordering.

\paragraph{Trade-offs and Bias.}
Intermixing success/failure examples (``mixed outcomes'') tends to raise
recall (e.g., ``label at start'') with a mild precision cost, consistent with
few-shot prompt biases reported in prior work on label/format sensitivity and
calibration. Importantly, absolute impacts are
small here, and overall PR-AUC remains stable across formats.

\begin{table*}[t]
\centering
\caption{Performance of BioBERT under different fine-tuning strategies. 
Top: in-distribution (seen interventions), middle: mixed-ratio experiment, 
bottom: out-of-distribution generalization (unseen interventions). 
Best results are \textbf{bolded}, second best are \underline{underlined}.}
\label{tab:combined_final}
\renewcommand{\arraystretch}{1.2}
\small
\begin{tabularx}{\linewidth}{
  >{\centering\arraybackslash}p{2.8cm}
  *{5}{>{\centering\arraybackslash}X}
}
\toprule
\textbf{Training Data} & \textbf{Accuracy} & \textbf{Precision} & \textbf{Recall} & \textbf{ROC-AUC} & \textbf{PR-AUC} \\
\midrule
\multicolumn{6}{c}{\textbf{(a) In-Distribution}} \\
\midrule
No Fine-Tuning & 0.527 ± 0.009 & 0.526 ± 0.008 & \textbf{0.937 ± 0.090} & 0.517 ± 0.035 & 0.531 ± 0.029 \\ 
Synthetic-Only & 0.534 ± 0.010 & 0.532 ± 0.005 & 0.873 ± 0.048 & 0.559 ± 0.009 & 0.584 ± 0.006 \\ 
Real-Only & \underline{0.633 ± 0.003} & \textbf{0.677 ± 0.006} & 0.563 ± 0.009 & \underline{0.689 ± 0.009} & \underline{0.723 ± 0.013} \\ 
Hybrid & \textbf{0.642 ± 0.012} & \underline{0.662 ± 0.023} & \underline{0.642 ± 0.027} & \textbf{0.698 ± 0.012} & \textbf{0.725 ± 0.016} \\ 
\midrule
\multicolumn{6}{c}{\textbf{(b) Ratio Mix}} \\
\midrule
100\% Synthetic & 0.545 ± 0.009 & 0.543 ± 0.010 & \textbf{0.825 ± 0.091} & 0.566 ± 0.015 & 0.589 ± 0.017 \\ 
80\% Syn + 20\% Real & 0.586 ± 0.015 & 0.594 ± 0.027 & 0.665 ± 0.068 & 0.620 ± 0.010 & 0.638 ± 0.011 \\ 
60\% Syn + 40\% Real & 0.608 ± 0.013 & 0.620 ± 0.005 & 0.640 ± 0.082 & 0.648 ± 0.019 & 0.667 ± 0.012 \\ 
40\% Syn + 60\% Real & \underline{0.620 ± 0.016} & 0.623 ± 0.018 & \underline{0.689 ± 0.092} & 0.670 ± 0.026 & 0.689 ± 0.028 \\ 
20\% Syn + 80\% Real & 0.629 ± 0.010 & \underline{0.642 ± 0.041} & 0.678 ± 0.091 & \underline{0.692 ± 0.008} & \underline{0.708 ± 0.001} \\ 
100\% Real & \textbf{0.638 ± 0.011} & \textbf{0.695 ± 0.002} & 0.542 ± 0.040 & \textbf{0.702 ± 0.009} & \textbf{0.728 ± 0.005} \\ 
\midrule
\multicolumn{6}{c}{\textbf{(c) Generalization}} \\
\midrule
No Fine-Tuning & 0.504 ± 0.005 & 0.502 ± 0.003 & \textbf{0.939 ± 0.086} & 0.512 ± 0.022 & 0.510 ± 0.020 \\ 
Synthetic-Only & 0.530 ± 0.005 & 0.518 ± 0.004 & \underline{0.883 ± 0.042} & 0.589 ± 0.020 & 0.574 ± 0.019 \\ 
Real-Only & \underline{0.645 ± 0.008} & 0.616 ± 0.007 & 0.771 ± 0.034 & \underline{0.709 ± 0.012} & \textbf{0.697 ± 0.010} \\ 
Hybrid & \textbf{0.658 ± 0.001} & \textbf{0.660 ± 0.006} & 0.652 ± 0.020 & \textbf{0.711 ± 0.001} & \underline{0.694 ± 0.004} \\ 
\bottomrule
\end{tabularx}
\end{table*}

\subsection{Error Analysis of Generated Trials}
While our main evaluation focused on aggregate performance, we also 
examined the generated trial rationales and narratives to better 
understand both their strengths and limitations.

\paragraph{Common Themes.} 
For \emph{successful trials}, the reasoning module frequently 
produced rationales referencing ``significant improvement in primary 
endpoints,'' ``absence of serious adverse events,'' or ``novel 
mechanisms proving effective.'' 
For \emph{failed trials}, typical rationales included ``lack of 
efficacy compared to control,'' ``unacceptable toxicity,'' or ``poor 
patient enrollment.'' 
These patterns are well aligned with known causes of clinical trial 
outcomes, providing reassurance of the generator’s plausibility.

\paragraph{Observed Artifacts.} 
We identified a few recurring issues: 
(i) the LLM occasionally attempted to insert conclusive statements 
such as ``Overall, the trial was a success,'' likely reflecting 
outcome phrases in the few-shot examples. To prevent label leakage, 
we removed such phrases during preprocessing. 
(ii) In some cases, the LLM generalized too broadly, e.g., 
``the drug improves survival'' without specifying effect size or 
statistics, or referenced elements like ``patient interviews,'' 
which are uncommon in trial registry reports. 
Importantly, these hallucinations did not undermine the correctness 
of the label or the basic logic of the trial narrative.

\paragraph{Implications.} 
This analysis clarifies why the synthetic data is useful: the model 
captures realistic outcome rationales and trial structures, while 
remaining largely free of critical errors. We acknowledge, however, 
that subtle inaccuracies remain. Future work may explore tighter 
constraints or verification steps to further improve fidelity and 
reduce hallucinations.

\subsection{Qualitative Audit of Synthetic Trials}

\subsubsection{Representative Exemplars}
Below we include short excerpts from several audited synthetic reports 
to illustrate their plausibility and structure. These are drawn directly 
from the generated set (with outcome labels stripped).

\begin{itemize}
    \item \textbf{Termination Reasons:}
    \begin{itemize}
        \item ``Insufficient efficacy and recruitment challenges'' 
        (synthetic\_clinical\_report\_1008.txt)
        \item ``Poor enrollment and high rates of treatment-related adverse events'' 
        (synthetic\_clinical\_report\_1009.txt)
        \item ``Insufficient patient accrual and low response rates'' 
        (synthetic\_clinical\_report\_3355.txt)
    \end{itemize}

    \item \textbf{Adverse Events:} gastrointestinal toxicity, rash, 
    and neutropenia (e.g., synthetic\_clinical\_report\_1003.txt, 
    \\
    synthetic\_clinical\_report\_1009.txt).

    \item \textbf{Balanced Efficacy/Safety:} 
    ``No serious adverse events were reported, with favorable tolerability 
    across patient cohorts'' (synthetic\_clinical\_report\_1807.txt).

    \item \textbf{Specific Endpoints:} 
    reports citing statistically significant improvement 
    with $p < 0.001$ in a ranolazine study 
    (synthetic\_clinical\_report\_1276.txt).
\end{itemize}

These exemplars demonstrate realistic trial narratives that mirror 
common patterns observed in real-world clinical studies.

\subsubsection{Audit Checklist}
We applied a simple checklist to each synthetic trial during manual review. 
A trial was considered \emph{plausible} if all items below were satisfied:

\begin{enumerate}
    \item \textbf{Objectives:} clearly stated study purpose or hypothesis.
    \item \textbf{Endpoints:} primary/secondary endpoints explicitly defined.
    \item \textbf{Adverse Events:} plausible side effects or safety findings reported.
    \item \textbf{Rationale:} outcome reasoning consistent with intervention and design.
    \item \textbf{Structure:} formatting follows ClinicalTrials.gov style 
    (title, arms, outcomes, termination reasons).
\end{enumerate}

This checklist ensured that generated trials were not only fluent, 
but also structurally and clinically consistent. Although we did not 
conduct a formal expert evaluation, this documented audit provides 
supporting evidence of clinical plausibility.

\end{document}